\documentclass[runningheads]{llncs}
\usepackage[T1]{fontenc}
\usepackage{graphicx}
\usepackage{color}
\usepackage[colorlinks=true, urlcolor=blue]{hyperref}
\usepackage{amsmath}
\usepackage{amssymb}
\usepackage{booktabs}
\usepackage{multirow}

\begin{document}
\title{AID: Agent Intent from Diffusion for Multi-Agent Informative Path Planning}
\titlerunning{\textbf{AID:} \textbf{A}gent \textbf{I}ntent from \textbf{D}iffusion for MAIPP}
\author{
Jeric Lew~\thanks{Corresponding Author} \and Yuhong Cao \and
Derek Ming Siang Tan \and
Guillaume Sartoretti$^{\star}$
}
\authorrunning{J. Lew et al.}

\institute{
Department of Mechanical Engineering, National University of Singapore, Singapore
\email{\{jericlew,caoyuhong,derektan\}@u.nus.edu, mpegas@nus.edu.sg}
}

\index{Lew, Jeric}
\index{Cao, Yuhong}
\index{Tan, Derek Ming Siang}
\index{Sartoretti, Guillaume}
\maketitle

\begin{abstract}
Information gathering in large-scale or time-critical scenarios (e.g., environmental monitoring, search and rescue) requires broad coverage within limited time budgets, motivating the use of multi-agent systems.
These scenarios are commonly formulated as multi-agent informative path planning (MAIPP), where multiple agents must coordinate to maximize information gain while operating under budget constraints.
A central challenge in MAIPP is ensuring effective coordination while the belief over the environment evolves with incoming measurements.
Recent learning-based approaches address this by using distributions over future positions as "intent" to support coordination.
However, these autoregressive intent predictors are computationally expensive and prone to compounding errors.
Inspired by the effectiveness of diffusion models as expressive, long-horizon policies, we propose \textit{AID}, a fully decentralized MAIPP framework that leverages diffusion models to generate long-term trajectories in a non-autoregressive manner. \textit{AID} first performs behavior cloning on trajectories produced by existing MAIPP planners and then fine-tunes the policy using reinforcement learning via Diffusion Policy Policy Optimization (DPPO). This two-stage pipeline enables the policy to inherit expert behavior while learning improved coordination through online reward feedback.
Experiments demonstrate that \textit{AID} consistently improves upon the MAIPP planners it is trained from, achieving \textbf{\boldmath $4\times$ faster} execution and up to \textbf{\boldmath $17\%$ increased information gain}, while scaling effectively to larger numbers of agents. Our implementation is publicly available at \href{https://github.com/marmotlab/AID}{github.com/marmotlab/AID}.

\end{abstract}

\section{Introduction}
Information gathering tasks, such as inspections of large-scale structures~\cite{inspection}, environmental monitoring~\cite{environment_monitor}, or search and rescue operations~\cite{search_n_rescue}, are often laborious, time-consuming, and potentially hazardous for humans. Autonomous robotic systems offer a safer and more efficient alternative that can operate in complex and dangerous environments with minimal human supervision. The difficulty in such scenarios lies in determining how robots should travel to efficiently collect information with minimal \textit{a priori} information, which has motivated extensive research in adaptive \textit{Informative Path Planning} (IPP)~\cite{akin_adaptive_2015,popovic_informative_2020}.

\begin{figure}[t]
  \centering
  \includegraphics[width=0.95\textwidth]{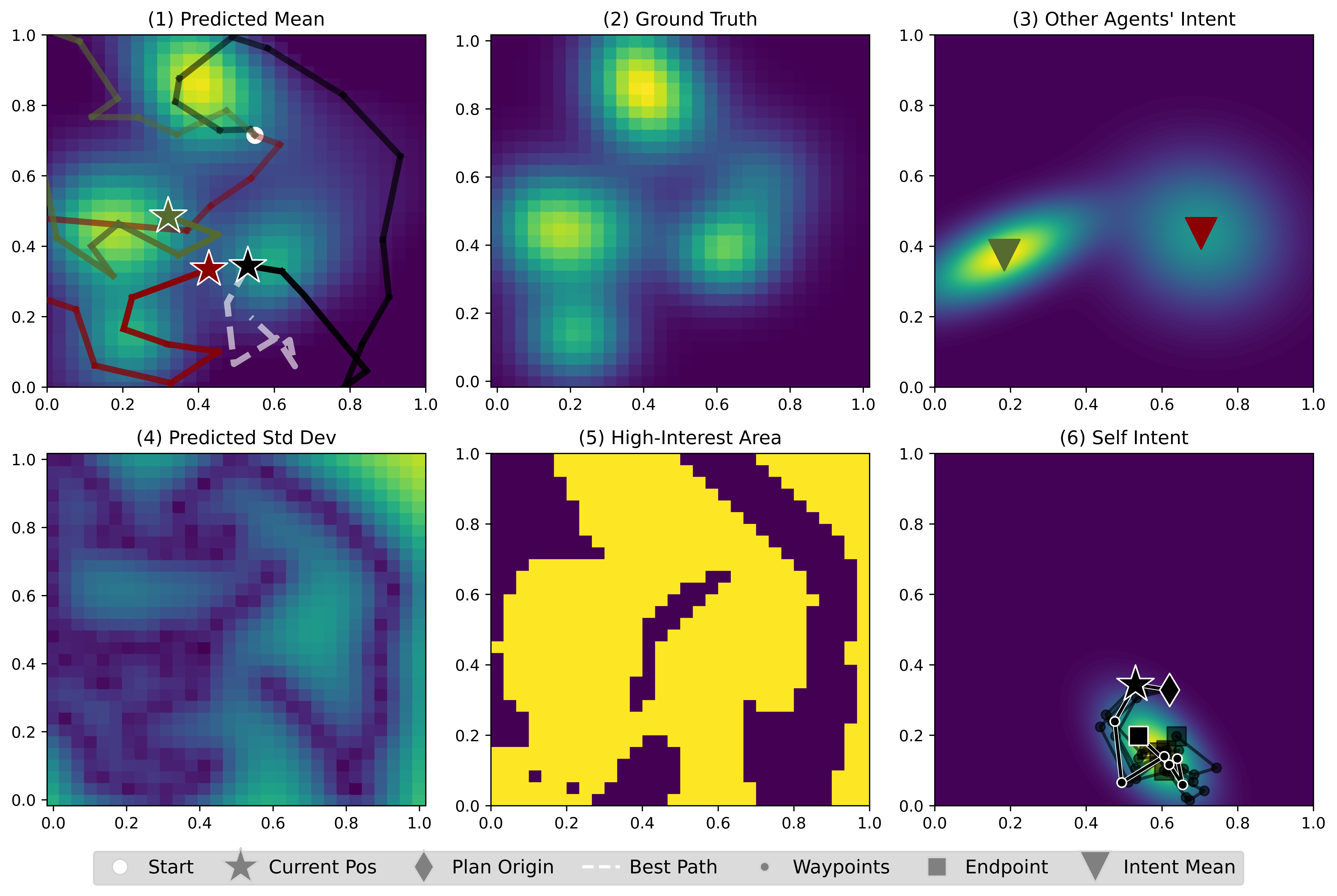}
  \caption[\textbf{Example run of \textit{AID} with 3 agents}]{\textbf{Example run of \textit{AID} with 3 agents.}
  \textbf{(1)} shows agent trajectories (colored lines) with the best planned path of the black agent.
  \textbf{(1)} and \textbf{(4)} depict the GP-predicted mean and standard deviation of the information distribution (Section~\ref{sec: gp}). 
  \textbf{(2)} shows the ground-truth information distribution, and \textbf{(5)} highlights the current high-interest area (Section~\ref{sec: maipp_problem}).
  \textbf{(6)} visualizes the black agent's intent distribution, while \textbf{(3)} shows the fused intent distribution of the other agents (Section~\ref{sec: intent}).}
  \label{fig: MAIPP Example}
\end{figure}

In adaptive IPP, an autonomous agent must plan a path that maximizes information gain about an environment while satisfying constraints such as distance or time. The environment is typically modeled as a continuous spatial field inferred from sparse, noisy measurements, with Gaussian Processes commonly used to maintain both predictions and uncertainty~\cite{cao2023catnipp,hitz_adaptive_2017,popovic_informative_2020,yang2023intentmaipp}. As the agent begins with little prior knowledge, each new measurement reshapes the estimated information distribution and shifts which regions are most informative, requiring frequent replanning and real-time computation. As IPP is NP-hard~\cite{krause2008optimizing}, existing solutions, ranging from meta-heuristics to deep reinforcement learning~\cite{cao2023catnipp,hitz_adaptive_2017,popovic_informative_2020,popo_rl}, aim to balance solution quality with computational efficiency.

Expanding upon single-agent IPP, \textit{multi-agent informative path planning} (MAIPP) aims to accelerate information gathering and expand coverage~\cite{MA_ipp,popo_ma}, which is particularly valuable in time-critical scenarios like post-disaster search and rescue. MAIPP has added challenges as agents must coordinate to maximize collective information gain, avoid redundant coverage, and adapt to a dynamically changing belief, making decentralized cooperation a difficult open problem.

MAIPP methods can rely on \textit{Sequential Greedy Assignment} (SGA)~\cite{corah2017efficient}, where agents plan one after another using single-agent IPP solvers while considering previously assigned paths to maintain coordination. Recent works, such as IntentMAIPP~\cite{yang2023intentmaipp}, adopt deep reinforcement learning (DRL) and models each agent’s future trajectory as "intent" to promote cooperative behavior. However, IntentMAIPP generates trajectories autoregressively which is computationally costly and results in noisy intent predictions when prediction horizon increases.

To address the challenges of MAIPP, the limitations of autoregressive methods and the need for long-horizon intent predictions, we propose \textit{AID} (\textbf{A}gent \textbf{I}ntent from \textbf{D}iffusion), a diffusion-based framework for MAIPP. Diffusion models, are well suited for planning and control because they can represent complex, multi-modal action distributions and naturally support long-term trajectory generation~\cite{cao2025dare,chi2024diffusionpolicy,janner2022diffuser}.
\textit{AID} leverages these properties by first cloning the behavior of existing MAIPP solvers and then fine-tuning via online reinforcement learning to produce faster and less noisy intent predictions. This approach improves multi-agent coordination and information gathering while maintaining computational efficiency. We demonstrate the flexibility and scalability of \textit{AID} by applying it to both sampling-based planners such as RIGTree~\cite{Hollinger_2014_ippsample} and DRL-based planners such as IntentMAIPP~\cite{yang2023intentmaipp}, showing consistent improvements, robust performance with increasing team size, and up to a $4\times$ speedup. Our results indicate that diffusion-based policies are an effective tool for MAIPP, facilitating long-horizon planning and decentralized cooperation through fine-tuning.

\section{Related Works}
\subsubsection{Single-Agent IPP:}
Single-agent IPP has been addressed through a variety of traditional planners, including sampling-based~\cite{hitz_adaptive_2017,Hollinger_2014_ippsample} methods and viewpoint-selection approaches~\cite{binney2012branchboundipp,meliou2007nonmyopic}. Among these, RIGTree~\cite{Hollinger_2014_ippsample} is a RRT-style sampling-based solver that select high information-gain branches efficiently, making it suitable for iterative replanning in adaptive IPP.
More recently, deep reinforcement learning~\cite{popovic2024learning,tan2025searchtta,wei2020ipprl} has emerged as strong alternative. CAtNIPP~\cite{cao2023catnipp} captures information of the environment through a probabilistic roadmap and uses a graph attention network to embed global belief information for local decision making. This attention mechanism enables fast, context-aware action selection, improving both solution quality and inference efficiency.

\subsubsection{Multi-Agent IPP:}
MAIPP remains relatively underexplored. An effective baseline is Sequential Greedy Assignment (SGA)~\cite{corah2017efficient}, where agents plan one after another while accounting for other agent's planned paths. SGA is simple, scalable, and compatible with any single-agent IPP planner, but frequent updates to the shared belief map often require repeated replanning, which can hinder coordinated behavior.
As in single-agent IPP, reinforcement learning has similarly proven to be an effective alternative in MAIPP~\cite{vashisth2025scalablemultirobotinformativepath,popo_ma,maipprl1}.
IntentMAIPP~\cite{yang2023intentmaipp}, building on CAtNIPP, plans using the global belief and an intent map that encodes other agents’ future positions as Gaussian distributions. This compact representation allows easy combination of multiple agents’ intents, but as more future nodes are sampled, compounding errors can accumulate, degrading long-term planning performance.

\subsubsection{Diffusion for Multi-Agent Systems:}
Diffusion models~\cite{ho2020denoisingdiffusionprobabilisticmodels,sohl2015deep} are a class of generative models that have achieved state-of-the-art performance in image synthesis. Their ability to capture multi-modal action distributions and suitability for high-dimensional output spaces have driven growing interest in robotics applications. Recent works have demonstrated the effectiveness of diffusion models in generating robot behaviors for manipulation~\cite{chi2024diffusionpolicy}, legged locomotion~\cite{huang2024diffuselocorealtimeleggedlocomotion}, and visual navigation~\cite{shah2023vint,Nomad}, among others. In multi-agent settings, however, the use of diffusion remains limited. Some existing approaches rely on offline datasets to learn trajectory distributions that ensure coordination or collision avoidance in multi-agent path finding~\cite{shaoul2024multi}. MADIFF~\cite{zhu2023madiff} introduces a multi-agent diffusion framework but incorporates centralized elements by performing attention across all agents at every decoder layer, which may limit scalability and decentralization. In contrast, our method extends the use of diffusion models toward decentralized multi-agent coordination by leveraging their long-horizon trajectory generation capabilities to model intent. The use of behavior cloning further aligns with a promising direction of distilling optimal solutions into scalable decentralized policies~\cite{Prorok2022aamas}.

\section{Background}
\subsection{Gaussian Processes (GPs)}  
\label{sec: gp}
In IPP, an agent's objective is to collect information from an environment where the underlying information is modeled as a continuous function over a 2D space, $\zeta: \mathcal{E} \rightarrow \mathbb{R}$, with $\mathcal{E} \subset \mathbb{R}^2$ representing the environment. However, this function is unknown to the agent(s) and must be inferred from limited measurements. Following previous works~\cite{cao2023catnipp,hitz_adaptive_2017,popovic_informative_2020,yang2023intentmaipp}, Gaussian Processes (GPs) provide a way to estimate this unknown function by interpolating between sparse observations.  

A GP defines a distribution over functions and enables us to approximate the true function $\zeta$ using a probabilistic model, $\zeta \approx GP(\mu, P)$, where $\mu$ and $P$ represent the mean and covariance functions of the GP, respectively. Given a set of $n$ measurement locations $X \subset \mathcal{E}$ and corresponding observations $Y$, along with a set of query locations $X^* \subset \mathcal{E}$ where the agent seeks to infer values, the GP posterior mean $\mu(X^*)$ and covariance $P(X^*)$ are given by:  

\begin{equation}
\mu(X^*) = \mu(X^*) + K(X^*, X)[K(X, X) + \sigma_n^2 I]^{-1} (Y - \mu(X)),
\end{equation}  
\begin{equation}
P(X^*) = K(X^*, X^*) - K(X^*, X)[K(X, X) + \sigma_n^2 I]^{-1} K(X^*, X)^T,
\end{equation}

where $K(\cdot, \cdot)$ is a kernel function defining the spatial correlation, $\sigma_n^2$ represents the measurement noise, and $I$ is the identity matrix. In this work, we utilize the Matérn 3/2 kernel, commonly used in IPP literature~\cite{cao2023catnipp,hitz_adaptive_2017,popovic_informative_2020,yang2023intentmaipp}.  

\subsection{Multi-Agent Informative Path Planning}
\label{sec: maipp_problem}
In MAIPP, the objective is to determine an optimal set of trajectories $\psi^* = \{\psi_1, \dots, \psi_m\}$ for $m$ agents, such that the collective information gain is maximized while ensuring that each agent adheres to its individual budget constraint. The problem is formulated as:  

\begin{equation}
\psi^* = \arg\max_{\psi \in \Psi} \sum_{i=1}^m I(\psi_i), \quad \text{s.t.} \quad C(\psi_i) \leq B, \quad 1 \leq i \leq m,
\end{equation}

where $I(\psi_i)$ represents the information gain from trajectory $\psi_i$, $C(\psi_i)$ denotes the trajectory cost, and $B$ is the allocated budget (path length) for each agent.  

Following prior works~\cite{cao2023catnipp,yang2023intentmaipp}, the information gain $I(\psi_i)$ is defined as the reduction in uncertainty over high-interest areas: $I(\psi_i) = \text{Tr}(P^-_I) - \text{Tr}(P^+_I)$, where $\text{Tr}(\cdot)$ denotes the trace of a matrix, and $P^-_I$ and $P^+_I$ represent the prior and posterior covariance matrices of the high-interest areas, respectively. These high-interest areas are determined using an upper confidence bound, $X_I = \{x_i \in X^* \mid \mu_i^- + \beta P^-_{i,i} \geq \mu_{th} \}$, where $\mu_i^-$ and $P^-_{i,i}$ correspond to the mean and variance of the GP at location $x_i$, while $\mu_{th}$ and $\beta$ control the threshold and confidence interval, respectively.

Thus, the planner must adapt to new measurements, as they continuously update the agent’s belief and redefine high-interest areas. This necessitates online replanning to maximize information gain and ensure efficient exploration. Figure~\ref{fig: MAIPP Example} illustrates the MAIPP problem, the use of GPs to estimate the information in the environment and an example run of \textit{AID}.

\begin{figure}[ht]
  \centering
  \includegraphics[width=0.65\textwidth]{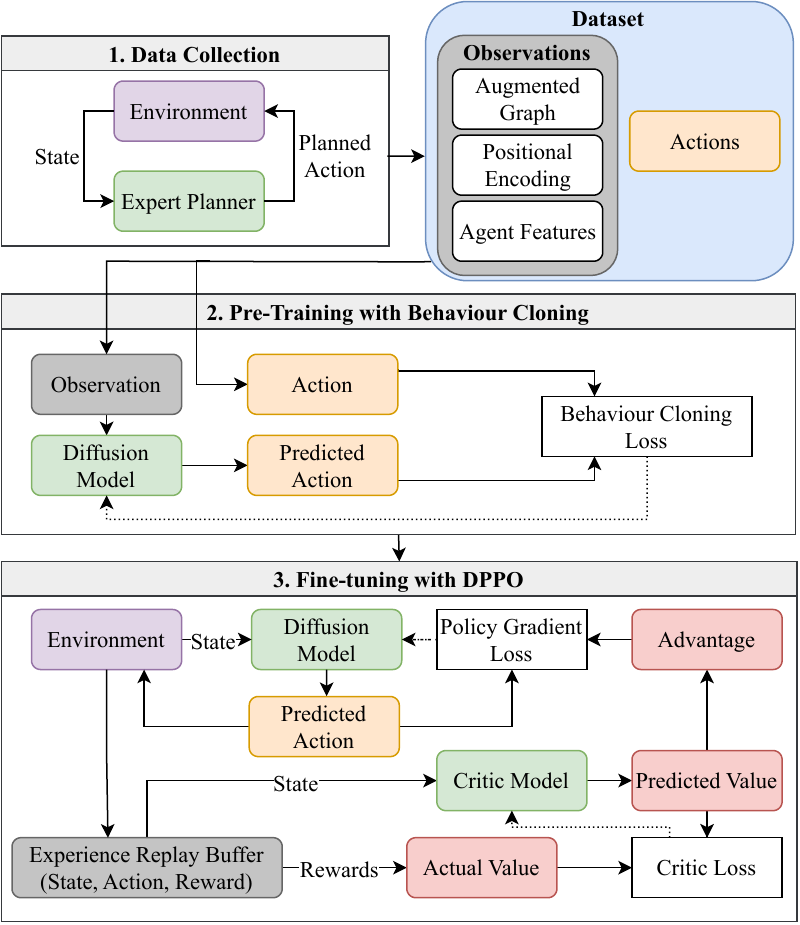}
  \caption[\textbf{Pipeline for \textit{AID}}]{\textbf{Pipeline for \textit{AID}.}}
  \label{fig: AID Pipeline}
\end{figure}

\section{Method}

To leverage diffusion models for MAIPP, our proposed \textit{AID} framework adopts a fully decentralized two-stage approach. In the first stage, we pre-train a diffusion policy via behavior cloning on a dataset of trajectories generated by existing MAIPP planners. This follows prior work on diffusion-based policies~\cite{chi2024diffusionpolicy,janner2022diffuser}.

To go beyond behavior cloning and further enhance agent coordination, we employ \textit{Diffusion Policy Policy Optimization} (DPPO)~\cite{dppo2024}, a recent reinforcement learning approach that fine-tunes pre-trained diffusion policies online to improve performance. By combining supervised pre-training with online fine-tuning, \textit{AID} achieves both improved sample efficiency and stronger policy performance in contrast to training a diffusion policy from scratch with DPPO.

This two-stage training pipeline allows \textit{AID} to fully exploit the advantages of diffusion models: representing complex action distributions and generating long-horizon trajectories in a single forward pass. Unlike autoregressive policies such as those used in IntentMAIPP~\cite{yang2023intentmaipp}, which require costly iterative updates and can accumulate compounding errors, \textit{AID} outputs temporally consistent plans, enabling more effective long-term planning and coordination across multiple agents. The overall pipeline of our framework is illustrated in Figure~\ref{fig: AID Pipeline}.

\subsection{Sequential Decision-Making Problem}
We formulate MAIPP as a sequential decision-making problem in continuous space, providing greater flexibility compared to previous works that employ DRL on a graph~\cite{cao2023catnipp,yang2023intentmaipp}. We assume an obstacle-free environment, but obstacle handling can be incorporated by adding standard collision-checking and avoidance.

Each agent $i$ starts from the same initial position and moves asynchronously to their next position which can be of different distance for each agent. Thus, the time steps, $t$, each agent can take before exhausting their budget might be different. Agents iteratively plan and execute their paths in a receding horizon manner until their budget is exhausted. At that point, the final trajectory of agent $i$ is given by: $\psi_i = \{\psi_1^i, \dots, \psi_t^i\}, \ \forall \psi_n^i \in \mathbb{R}^2$
with trajectory length $C(\psi_i) = \sum_{j=1}^{n_i-1} L_2(\psi_j^i, \psi_{j+1}^i)$, where $L_2(\cdot, \cdot)$ denotes the Euclidean distance.

\begin{figure}
    \centering
    \includegraphics[width=\textwidth]{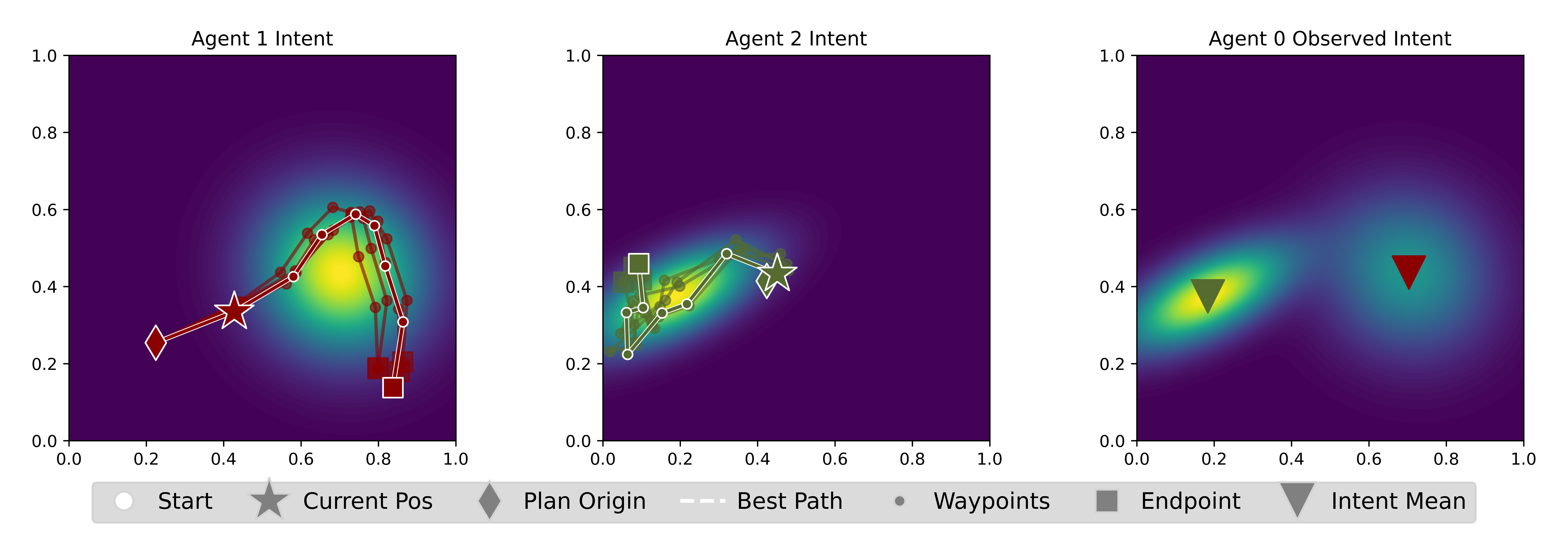}
    \caption[\textbf{Agent intent generated by diffusion model}]{
    \textbf{Agent intent generated by diffusion model.} 5 trajectory predictions were generated per agent with a planning horizon $T_p = 8$ (Section~\ref{sec: pretrain}).
    Elements with white borders denote the chosen path that is executed.
    The fused intent visualization shows the normalized distribution of positions from other agents' sampled trajectories.
    }
    \label{fig: intent}
\end{figure}

\subsection{Modeling Agent's Intent}
\label{sec: intent}
To facilitate coordination among agents, we represent each agent’s planned future positions as a probabilistic distribution, termed as \textit{intent}~\cite{yang2023intentmaipp}. Each agent updates and shares its intent when it plans a new trajectory, allowing others to incorporate this information into their decision-making. 

An agent's intent is modeled as a Gaussian distribution $GD(\mu^i(t), \Sigma^i(t))$, where $\mu^i(t)$ and $\Sigma^i(t)$ are the mean and covariance matrix fitted to the planned trajectory of agent $i$ at time step $t$. This distribution provides a probabilistic estimate of where the agent is likely to move.

To ensure effective coordination, each agent aggregates the intent of all other $m-1$ agents to form a fused intent map. This is achieved by summing the Gaussian distributions of the other agents' intents and normalizing the result. The fused intent map, shown in Figure~\ref{fig: intent}, serves as an additional input to the agent’s decision-making process, allowing it to make more informed path-planning decisions while adapting to the predicted movements of its teammates.

\subsection{Pre-training with Behavior Cloning}
\label{sec: pretrain}
In the first stage of \textit{AID}, we pre-train a diffusion policy using behavior cloning. This stage aims to initialize the diffusion model with behaviors from existing MAIPP planners before fine-tuning it through reinforcement learning. To achieve this, we construct a dataset of trajectories generated by existing MAIPP planners. Each agent’s state-action pairs are recorded independently at every time step, following the same input–output structure used by the diffusion policy. The collected dataset thus consists of pairs of \textit{observations} and corresponding \textit{actions}, defined as follows:

\subsubsection{Observation:} The input (observation) to the diffusion model for agent $i$ at time step $t$ is $s^i_t = \{V^{'i}, Q^i, A^i\}$. The augmented graph nodes, $V^{'i}$, represents information in the environment and is derived from a probabilistic roadmap (PRM), where there are $n$ nodes which are connected to their $k$ nearest neighbors, forming $G^i = (V^i, E^i), \ i \in \{1, ..., m\}$, where $V^i = \{v^i_1, ..., v^i_n\}$ represents the nodes and $E^i$ the edges. The PRM graph is then augmented with additional information such that each node $v^{'i}_j = (\hat{x}^i_j, \hat{y}^i_j, \mu(v^i_j), P(v^i_j), f(v^i_j)) \in V^{'i}, \ j \in \{1, ..., n\}$, where $(\hat{x}, \hat{y})$ is the relative position from the agent's current location, $\mu(v^i_j), P(v^i_j)$ are the mean and variance of the predicted information from the GP, and $f(v^i_j)$ is the intent information of the other $m-1$ agents at the current node's position. The positional encoding of the graph, $Q^i$, is computed based on graph connectivity using eigenvalues~\cite{dwivedi2021generalization}, resulting in a 32-dimensional representation per node. Lastly, the agent-specific input $A^i$ is defined as $A^i = (v^i_t, B^i_t, \mu_{th})$, where $v^i_t$ is the agent’s current position, $B^i_t = B - C(\psi^i)$ represents the remaining budget, and $\mu_{th}$ is the threshold for defining high-interest areas.

\subsubsection{Action:} At each time step $t$, agent $i$ processes its updated observation $s^i_t$ through a diffusion model, which outputs a sequence of delta $x$ and $y$ positions $\Delta^i_t = \{\delta^i_t, \delta^i_{t+1}, ..., \delta^i_{t+T_p-1}\}$ over a horizon of length $T_p$. This sequence of delta positions are then cumulatively summed to yield a sequence of future way-points $\psi^i_{t+1:t+T_p} = \{\psi^i_{t+1}, \psi^i_{t+2}, ..., \psi^i_{t+T_p}\}, \ \psi^i_j \in \mathbb{R}^2$. This represents the predicted path, which is executed in a receding horizon manner over $T_a$ steps to maintain adaptability in response to new information from updated observations.

As the data is collected offline, we have access to complete trajectories, providing flexibility in defining how intent information is derived. If intent is available from the existing MAIPP planner (e.g. in sampling-based methods where a planned path is provided), we use it directly. Otherwise, intent can be retroactively assigned based on the agent's actual future positions. The same approach is applied when collecting the actions from offline trajectories. 
With the collected trajectories, we pre-train the diffusion policy following Chi et al.~\cite{chi2024diffusionpolicy} using a behavior cloning objective. Specifically, for each agent $i$ at time step $t$, Gaussian noise $\varepsilon_k \sim \mathcal{N}(0, I)$ is added to the ground-truth delta action sequence $\Delta_t^{0,i}$, and the diffusion model $\varepsilon_\theta$ is trained to predict this noise conditioned on the agent’s observation $s_t^i$ and the noisy action sequence at diffusion step $k$. The resulting mean squared error loss is:

\begin{equation}
\mathcal{L}_{\text{BC}} = \text{MSE}\Big(\varepsilon_k, \ \varepsilon_\theta(s_t^i, \ \Delta_t^{0,i} + \varepsilon_k, \ k)\Big),
\end{equation}

\subsection{Fine-Tuning with DPPO}
After pre-training the diffusion policy via behavior cloning, we fine-tune it online using reinforcement learning. The policy interacts with the environment to collect rollout trajectories containing observations, actions, rewards, and value estimates, which are then used for optimization.

DPPO (Diffusion Proximal Policy Optimization)~\cite{dppo2024} extends PPO to diffusion policies by treating each denoising step as an action within a \textit{Diffusion MDP}. This allows reward signals to propagate through the entire denoising process, enabling end-to-end policy improvement. As in PPO, DPPO stabilizes training using a clipped surrogate objective.

A critic network is trained alongside the policy to predict state values, which are used to compute advantage estimates via Generalized Advantage Estimation (GAE)~\cite{DBLP:journals/corr/SchulmanMLJA15}. These advantage estimates guide the policy to favor denoising actions that yield higher cumulative rewards. The critic itself is optimized using a MSE loss between predicted state values and the returns computed with GAE, ensuring accurate value estimation and improving the quality of policy updates during fine-tuning.

\subsubsection{Reward:}
When an agent depletes its budget, it receives a negative reward:

\begin{equation}
    r_f = -\alpha \left( \frac{\text{Tr}(P^f)}{\text{Tr}(P^i)} \right)^\beta,
\end{equation}

where $\text{Tr}(P^i)$ and $\text{Tr}(P^f)$ denote the initial and final covariance traces over high-interest regions, $\alpha$ is a scaling factor, and $\beta\in[0,1]$ controls the concavity of the penalty ($\alpha=5$, $\beta=0.5$ in practice). A concave function is chosen because reducing uncertainty becomes increasingly difficult as the covariance trace approaches zero. With $\beta<1$, the reward structure reflects this by assigning relatively greater incentive to small but harder reductions at low uncertainty, compared to easy reductions of the same magnitude when uncertainty is high. This sparse reward directly encourages agents to minimize final uncertainty~\cite{yang2023intentmaipp}, promoting cooperative information gathering during fine-tuning.

\subsection{Neural Network Architecture}
\subsubsection{Graph Attention Encoder:}
Given input observation $s_t^i$, each feature in $s_t^i$, namely $\{V^{'i}, Q^i, A^i\}$, is projected into a $d$-dimensional embedding, $\{\hat{V^{'i}}, \hat{Q^i}, \hat{A^i}\}$. The embedded node features, $\hat{V^{'i}}$, are summed with the corresponding embedded positional encoding, $\hat{Q^i}$, and passed through a \textit{multi-head self-attention}~\cite{vaswani2017attention} layer, following the approach in~\cite{cao2023catnipp,yang2023intentmaipp}, to capture relational dependencies between different nodes in the environment. The output is a set of enhanced node features, which are then passed through a \textit{multi-head cross-attention} layer, where the enhanced node features serve as the key and value, while the agent's embedded features, $\hat{A^i}$, serve as the query. This cross-attention mechanism allows the encoder to focus on the most relevant parts of the environment, taking into account the agent's current state, such as its location and remaining budget. The final output is a $d$-dimensional encoded state representation, $\hat{s_t^i}$, which encapsulates agent $i$'s perception of the environment. This representation is then used as the input for the diffusion model and critic model, with each network having its own graph attention encoder.

\subsubsection{Diffusion Model:}
Conditioned on $\hat{s_t^i}$, a U-Net-based diffusion model~\cite{chi2024diffusionpolicy} iteratively denoises a randomly sampled action sequence of length $T_p$, producing a refined planned action sequence $\Delta^i_t = \{\delta^i_t, \delta^i_{t+1}, ..., \delta^i_{t+T_p-1}\}$.

\subsubsection{Critic Model:}
The critic takes $\hat{s_t^i}$ as input and passes it through a three-layer multi-layer perceptron (MLP) to predict the state value, $V(s_t^i)$, providing the necessary estimates for advantage computation during DPPO training.

\subsection{Implementation Details}
\subsubsection{Environment Setup:}
Training environments are generated by sampling between 8--12 Gaussian functions to produce diverse, multimodal belief landscapes. Each environment contains three agents ($m=3$). A Probabilistic Roadmap (PRM) with 200 nodes and $k = 20$ nearest neighbors is used. The high-interest threshold is set to $\mu_{th} = 0.4$, and the confidence parameter is $\beta = 1$. Each agent is allocated a budget of $B = 3$ steps per episode.

\subsubsection{Diffusion Policy:}
The diffusion model uses 20 denoising steps, with only the last 10 denoising steps fine-tuned with DPPO. The graph-attention encoder has an embedding dimension of $d = 64$ and processes two consecutive states from time steps $t$ and $t\!-\!1$. The planning horizon is $T_p = 8$ and the action horizon is $T_a = 2$. During fine-tuning and execution, the policy generates 5 candidate paths per step and executes the trajectory that yields the greatest reduction in covariance trace. These generated trajectories are also used to construct agent intent during multi-agent planning. Model training was done on an NVIDIA GeForce RTX 4080 SUPER. Diffusion policies were pre-trained for 1,000 epochs (9 hours), with the best-performing model selected via evaluation. Fine-tuning with DPPO was performed for up to 500 epochs or until performance plateaued. Total training time varied between 15–30 hours depending on the base planner and DPPO parameters. Additional details and code are available at \href{https://github.com/marmotlab/AID}{github.com/marmotlab/AID}.

\begin{table}[t]
\centering
\caption{Comparison of coverage and planning across 100 environments.}
\resizebox{\textwidth}{!}{
\begin{tabular}{l|cc|cc|cc}
\toprule
\multirow{2}{*}{Method} & 
\multicolumn{2}{c|}{3 Agents} & 
\multicolumn{2}{c|}{5 Agents} & 
\multicolumn{2}{c}{10 Agents} \\
& \textbf{Cov. Trace} & \textbf{Time (s)} &
  \textbf{Cov. Trace} & \textbf{Time (s)} &
  \textbf{Cov. Trace} & \textbf{Time (s)} \\
\midrule
RIGTreeSGA          & 68.4 $\pm$ 22.0 & 47.1 $\pm$ 1.1  & 22.7 $\pm$ 10.5 & 86.1 $\pm$ 1.2  & 4.5 $\pm$ 2.6 & 177.0 $\pm$ 4.0  \\
AID (PT RIGTreeSGA)       & 89.7 $\pm$ 27.4 & 9.5  $\pm$ 0.8  & 47.6 $\pm$ 23.3 & 16.9 $\pm$ 1.0  & 22.5 $\pm$ 12.8 & 32.2 $\pm$ 0.8  \\
AID (PT RIGTreeSGA + FT)       & 56.5 $\pm$ 15.2 & 11.0 $\pm$ 0.9  & 20.9 $\pm$ 8.9 & 19.6 $\pm$ 1.3  & 5.9 $\pm$ 4.0  & 38.2 $\pm$ 2.8  \\
\midrule
IntentMAIPP      & 28.5 $\pm$ 10.7 & 38.6 $\pm$ 2.0  & 8.7 $\pm$ 3.1  & 96.0 $\pm$ 5.2 & 3.4  $\pm$ 1.5  & 312.2 $\pm$ 2.8  \\
AID (PT IntentMAIPP)   & 33.5 $\pm$ 10.5 & 7.8  $\pm$ 0.9  & 12.0 $\pm$ 4.0 & 14.8 $\pm$ 1.3  & 5.0 $\pm$ 3.0  & 30.9 $\pm$ 1.5  \\
AID (PT IntentMAIPP + FT)   & 27.5 $\pm$ 6.6  & 13.0 $\pm$ 1.3  & 8.7 $\pm$ 3.6  & 27.3 $\pm$ 2.3 & 2.6  $\pm$ 0.9  & 67.3 $\pm$ 5.3  \\
\bottomrule
\end{tabular}
}
\label{tab:aid_results}
\end{table}

\section{Experiments}
To assess the effectiveness of \textit{AID}, we evaluate it on two representative MAIPP baselines. The first is RIGTreeSGA, an extension of RIGTree~\cite{Hollinger_2014_ippsample}, an RRT-based single-agent IPP planner, augmented with SGA~\cite{corah2017efficient} to enable multi-agent coordination. The second is IntentMAIPP~\cite{yang2023intentmaipp}, an attention-based reinforcement learning approach that autoregressively samples future intent trajectories.

All methods are evaluated on a fixed set of 100 unseen environments. Performance is measured using (i) the remaining covariance trace within high-interest regions which reflects the residual uncertainty over the environment’s underlying information, as defined in Section~\ref{sec: maipp_problem}, and (ii) the overall planning time, defined as the total time required for a planner to expend its budget, which captures computational efficiency. \textbf{PT} denotes diffusion models trained solely by behavior cloning their respective baseline planner, whereas \textbf{FT} denotes the same models after additional online fine-tuning.

As shown in Table~\ref{tab:aid_results}, diffusion-based policies consistently reduce planning time across all baselines and team sizes. \textit{AID} achieves approximately a \textbf{4$\times$ speedup} over sampling-based RIGTreeSGA and autoregressive IntentMAIPP, with the advantage increasing as the number of agents grows, highlighting its computational scalability.

In terms of information-gathering performance (covariance trace), the improvements are more nuanced. Pure behavior cloning (PT) variants exhibit higher variance and reduced performance compared to their experts, consistent with prior observations that diffusion policies are unable to fully match long-horizon expert behavior~\cite{chi2024diffusionpolicy,dppo2024}. However, after fine-tuning, performance becomes more competitive. With 3 agents, \textit{AID} (PT RIGTreeSGA + FT) improves covariance trace by approximately \textbf{17\%} over RIGTreeSGA while maintaining a significant runtime reduction. For IntentMAIPP, gains from fine-tuning are more modest, likely because the expert is already near-optimal, but comparable performance is achieved with substantially reduced runtime.

Finally, scaling to 5 and 10 agents shows similar trends: diffusion-based policies match or improve baseline performance within variance while planning significantly faster. These results suggest that AID primarily offers a favorable trade-off between information gain and computational efficiency, making it a scalable and practical approach for larger multi-agent teams.

\section{Conclusion}
In this work, we introduced \textit{AID}, a decentralized framework that integrates diffusion models into multi-agent informative path planning. Instead of relying on autoregressive policy sampling, diffusion models allow direct generation of long-horizon trajectories in a single forward pass, which supports coherent planning and provides a stable intent signal for coordination.

Through experiments on both sampling-based and learning-based MAIPP solvers, we showed that \textit{AID} improves multi-agent cooperation, increases information gain, and planning time when compared to existing methods. These results highlight the benefits of combining behavior cloning with online fine-tuning through DPPO to obtain robust diffusion policies for multi-agent planning.

Future work include scaling this algorithm to hundreds or even thousands of agents and applying \textit{AID} to 3D IPP environments with real robot experiments. We are also interested in extending \textit{AID} to other multi-agent tasks that require coordinated actions, like multi-drone cooperative manipulation.

\begin{credits}

\subsubsection{\ackname} This work was supported by NUS under grant TL/FS/2025/01.
\subsubsection{\discintname} The authors declare no conflicts of interest with respect to the research, authorship, and/or publication of this article
\end{credits}

\newpage
\bibliographystyle{splncs04}
\bibliography{ref}

\end{document}